\documentclass[final]{cvpr}

\usepackage[pagebackref=true,breaklinks=true,colorlinks,bookmarks=false]{hyperref}

\usepackage{times}
\usepackage{graphicx}
\usepackage{amsmath,amssymb,amsthm,mathabx}
\usepackage{booktabs}
\usepackage{algorithmic}
\usepackage[linesnumbered,ruled,vlined]{algorithm2e}
\usepackage{acronym}
\usepackage{enumitem}
\usepackage{setspace}
\usepackage[skip=3pt,font=small]{subcaption}
\usepackage[dvipsnames]{xcolor}
\usepackage[capitalise]{cleveref}
\usepackage{tabularx,multirow,array,makecell}

\DeclareMathOperator*{\argmax}{arg\,max}

\newcommand{\energy}{\mathcal{E}}

\makeatletter
\renewcommand{\paragraph}{%
  \@startsection{paragraph}{4}%
  {\z@}{0ex \@plus 0ex \@minus 0ex}{-1em}%
  {\hskip\parindent\normalfont\normalsize\bfseries}%
}
\makeatother

\graphicspath{{../figures/}}

\crefname{algocf}{Alg.}{Algs.}
\Crefname{algocf}{Algorithm}{Algorithm}

\definecolor{gblue}{HTML}{4285F4}
\definecolor{gred}{HTML}{DB4437}

\acrodef{tom}[ToM]{Theory of Mind}
\acrodef{pomdp}[POMDP]{partially observable Markov decision process}
\acrodef{map}[MAP]{maximum a posteriori}
\acrodef{mle}[MLE]{maximum likelihood estimation}

\usepackage{bbm,bm}
\usepackage{epigraph}
\usepackage{colortbl}

\definecolor{mygray}{gray}{.92}

\newcolumntype{a}{>{\columncolor{mygray}}c}
\newcolumntype{b}{>{\columncolor{white}}c}
\newcommand{\thickhline}{%
    \noalign {\ifnum 0=`}\fi \hrule height 1pt
    \futurelet \reserved@a \@xhline
}

\def\cvprPaperID{4411} %
\def\confYear{CVPR 2021}

\begin{document}

\title{Learning Triadic Belief Dynamics in Nonverbal Communication from Videos\vspace{-12pt}}

\author{
Lifeng Fan$\thanks{Lifeng Fan and Shuwen Qiu contributed equally.}$~,\hspace{1pt}
Shuwen Qiu$^{*}$,\hspace{1pt}
Zilong Zheng,\hspace{1pt}
Tao Gao,\hspace{1pt}
Song-Chun Zhu,\hspace{1pt}
Yixin Zhu\vspace{2pt}\\
UCLA Center for Vision, Cognition, Learning, and Autonomy\\
{\tt\small \{lfan, s.qiu, z.zheng\}@ucla.edu, \{tao.gao, sczhu\}@stat.ucla.edu, yixin.zhu@ucla.edu}\\
{\tt\small \url{https://github.com/LifengFan/Triadic-Belief-Dynamics}}
\vspace{-9pt}
}

\maketitle
\thispagestyle{empty}
\pagestyle{empty}

\begin{abstract}
Humans possess a unique social cognition capability~\cite{saxe2006uniquely,hinde1974biological}; nonverbal communication can convey rich social information among agents. In contrast, such crucial social characteristics are mostly missing in the existing scene understanding literature. In this paper, we incorporate different nonverbal communication cues (\eg, gaze, human poses, and gestures) to represent, model, learn, and infer agents' mental states from pure visual inputs. Crucially, such a mental representation takes the agent's belief into account so that it represents what the true world state is and infers the beliefs in each agent's mental state, which may differ from the true world states. By aggregating different beliefs and true world states, our model essentially forms ``five minds'' during the interactions between two agents. This ``five minds'' model differs from prior works that infer beliefs in an infinite recursion; instead, agents' beliefs are converged into a ``common mind''~\cite{levesque1990acting,tomasello2010origins}. Based on this representation, we further devise a hierarchical energy-based model that jointly tracks and predicts \textit{all} five minds. From this new perspective, a social event is interpreted by a series of nonverbal communication and belief dynamics, which transcends the classic keyframe video summary. In the experiments, we demonstrate that using such a social account provides a better video summary on videos with rich social interactions compared with state-of-the-art keyframe video summary methods.
\end{abstract}

\begin{figure*}[t!]
    \centering
    \vspace{-9pt}
    \includegraphics[width=\linewidth]{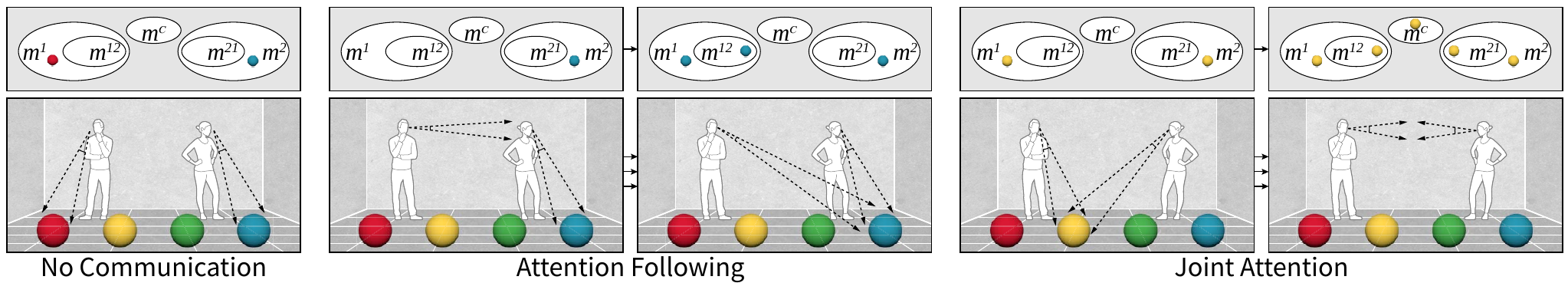}
    \caption{\textbf{Triadic belief dynamics in nonverbal communication.} Three types of communication events emerge from social interactions (bottom) and causally construct agents' belief dynamics (top). In this paper, we propose a novel structural mind representation ``five minds'' and a learning and inference algorithm for belief dynamics based on a hierarchical energy-based model that tracks (i) each agent's mental state ($m^1$ and $m^2$), (ii) their estimated belief about other agent's mental state ($m^{12}$ and $m^{21}$), and (iii) the common mind ($m^c$). Of note, some events have two phases connected by three arrows. }
    \label{fig:concept}
\end{figure*}

\vspace{-6pt}

\section{Introduction}

\epigraph{``The human body is the best picture of the human soul."}{--- \textup{Ludwig Wittgenstein}~\cite{wittgenstein1953philosophical}}

We live in a world with a plethora of animate and goal-directed agents~\cite{zhu2020dark}, or at least it is how humans perceive and \textit{construct}~\cite{white1990ideas} the world in our \textit{mental state}~\cite{hume1938abstract}. The iconic Heider-Simmel display~\cite{heider1944experimental} is a quintessential stimulus, wherein human participants are given videos of simple shapes roaming around the space. In this experiment, humans have a strong inclination to interpret the observed featureless motions composed of simple shapes as a story-telling description, such as a hero saving a victim from a bully. This social cognition account of human vision is largely missing in the computational literature of scene understanding or, more broadly, the field of computer vision.

In the field of social cognition, researchers have identified two unique components that distinguish human adults from infants and other primates~\cite{saxe2006uniquely}. The first component is ``\textbf{representational \ac{tom}},'' the ability to attribute mental states to oneself and others, to understand that others have perspectives and mental states different from one's own, as well as using these abilities to recognize false belief~\cite{premack1978does}. In the theoretical construct of mental states, mainstream psychology and related disciplines have traditionally treated \textit{belief} as one simplest form, and therefore one of the building blocks of conscious thought~\cite{horgan1992review}. Belief can be constructed as mental objects with semantic attributes; cognitive states and processes are constituted by the occurrence, transformation, and storage of such information-bearing structure~\cite{edward2020mental}. The second component is the \textbf{triadic relations}: \textit{You}, and \textit{Me}, collaboratively looking at, working on, or talking about \textit{This}~\cite{tomasello2010origins}. Much power of human social cognition depends on the ability to form representations with a triadic structure~\cite{saxe2006uniquely}.

To promote social cognition in computer vision, we focus on belief dynamics in nonverbal communication. Here, \textit{belief} is defined as an entity and its attributes (\eg, location), and \textit{belief dynamics} (\ie, the change of belief) are naturally and completely summarized using four categories: \textit{occur} indicates an agent becomes aware of an object at a certain location, \textit{update} means an agent knows the object's attribute was updated, \textit{disappear} denotes that an agent loses track of the object's attribute, and \textit{null} is no change. We emphasize on \textit{triadic relations} emerged during nonverbal communication, including \textit{No Communication}, \textit{Attention Following}, and \textit{Joint Attention}~\cite{fan2019understanding,admoni2017social}: \textit{No Communication} indicates no social interaction between the two agents, \textit{Attention Following} is a one-way observation, and \textit{Joint Attention} means that two agents have the same intention to share attention on a common stimulus and both know that they are sharing the attention~\cite{tomasello2010origins}; see an illustration in \cref{fig:concept}.

\setstretch{0.995}

To account for the two social components computationally, we propose a novel structural mind representation, termed ``\textbf{five minds},'' that includes two first-order self mental states (\ie, the ground-truth mental state), two second-order estimated mental states of each other's mind (may deviate from the ground-truth mental states), and the third-level ``common mind.'' Note that the proposed ``five minds'' differs from prior models that attempt to infer mental states among agents \textit{recursively} with potentially infinite loops; instead, the ``common mind'' considers what the two agents share completely transparently without infinite recursion and corresponds to the concept of ``common ground''~\cite{tomasello2010origins}.

The proposed ``five minds'' model is well-grounded to visual inputs, especially in terms of nonverbal communication. For instance, gaze communication uses eye gazes as portals inward to provide agents with glimpses into the inner mental world~\cite{fan2019understanding}, and pointing gesture serves as ``the first uniquely human forms of communication'' to ground and reshape mental states~\cite{tomasello2010origins}. We bring these crucial social components into representing, modeling, learning, and inference of belief dynamics in the computer vision community. Intuitively, the spatiotemporal parsing of social interactions affords the emergence of communication events; these events causally affect belief dynamics. Thus, a hierarchical energy-based model with Bayesian inference is naturally derived to track, maintain, and predict the mental states of all ``five minds.'' To demonstrate the model's efficacy, we collect a new 3D video dataset with eye-tracking devices to facilitate ground-truth labeling. We verify the proposed method on this new 3D video dataset focusing on rich nonverbal social interactions and triadic belief dynamics.

This paper makes four contributions:
(i) By incorporating crucial social cognition components, we address a new task of triadic belief dynamics learning and inference from nonverbal communication in natural scenes with rich social interactions. We propose a novel structural mental representation ``five minds'' by introducing a ``common mind,'' with well-defined and quantized belief and belief dynamics, as well as nonverbal communication events. To the best of our knowledge, ours is the first to tackle such challenging problems in the field of computer vision.
(ii) We collect a new 3D video dataset with rich social interactions using eye-tracking devices to facilitate ground-truth labeling; nonverbal communication events and belief dynamics are densely annotated. Such a setup goes beyond toy and symbolic examples presented in the literature, which we believe will serve as a modern benchmark for high-level social learning based on pixel inputs.
(iii) We devise a hierarchical energy-based model and a beam-search-based algorithm to simultaneously optimize the learning and inference of nonverbal communication events and belief dynamics.
(iv) We provide a benchmark and demonstrate the efficacy of the proposed method in a keyframe-based video summary.

\section{Related Work}

\paragraph{Nonverbal behavior and human communication}

Tomasello~\cite{tomasello2010origins} argues that nonverbal communication is the ``unconventionalized and uncoded'' form, more foundational than the human natural language. Crucially, instead of merely treating head and body motions as an assembly of skeletons movements (\eg, gaze~\cite{kellnhofer2019gaze360}, gesture~\cite{narayana2018gesture}, or interaction~\cite{joo2019towards} in computer vision), we do recognize the underlying intentions behind these motions from the perspective of human social cognition; pointing and iconic gestures have their special meaning to convey the message and establish shared intentionality and common ground~\cite{gilbert1992social}.

This unique view of nonverbal behavior and communication is largely ignored in modern scene understanding and computer vision. The present work subsumes prior work in gaze, gesture, body motions, and interactions in computer vision by presenting a hierarchical graphical representation, wherein the communication events~\cite{fan2019understanding,fan2018inferring} emerge from the spatiotemporal parsing of low-level signals to maintain the triadic relations and belief dynamics among agents.

\paragraph{Machine \ac{tom}}

\ac{tom} has been long regarded as an acid test for human social interaction; impairment of such capability to construe persons in terms of their inner mental lives often results in autism~\cite{baron1995children}. In literature, modern computational models of \ac{tom} often treat the inference of mental states as \textit{infinite} (or approximated by finite) recursions, notably by \ac{pomdp}~\cite{han2018learning,baker2011bayesian,doshi2010modeling,de2013much}. Recent research includes estimating the opponent's sophistication (\ie, recursion) in sequential games~\cite{wunder2011using,de2017estimating}, representing and updating beliefs through time~\cite{zettlemoyer2009multi,panella2017interactive}, reasoning about other agent's desires and beliefs based on their actions with a Bayesian account~\cite{baker2009action}, or learning to model agent's mental state and policy in grid world~\cite{rabinowitz2018machine}. However, studies have concluded that the default level of recursive reasoning typically could go no deeper than one or two levels~\cite{camerer2004cognitive}; instead, we tend to build and rely on the ``common mind''~\cite{tang2020bootstrapping,stacy2020intuitive} after only one or two levels of recursive reasoning of mental states.

In this paper, we adopt the representation of ``common mind,'' a crucial ingredient to properly interpret the triadic relation \textit{without} infinite recursion. Sharing a similar spirit, the coordinated and joint planning~\cite{grosz1996collaborative,grosz2006dynamics,kleiman2016coordinate,ho2016feature} has been extensively studied in symbolic-like environments. Additional efforts have also emerged to recognize false-belief or perspective-taking with more realistic and noisy data in the field of robotics~\cite{yuan2020joint,milliez2014framework}, computer vision~\cite{eysenbach2016mistaken}, and natural language processing~\cite{nematzadeh2018evaluating}. However, the problem settings in prior work either lack rich interactions or communications among agents or have relatively confined problem space (at most on object/human tracking). In comparison, the problem setting in this paper considers rich social interactions with nonverbal communication in physical indoor environments captured and synced with multiple Azure Kinect sensors and eye-tracking devices. To tackle the challenges introduced by raw video input, we present a much more expressive hierarchical representation to interpret the interactions and communications among agents.

\paragraph{Keyframe-based video summary}

Keyframe-based video summary is a practical application of video understanding. In literature, models tend to obtain keyshots for segment-based summary~\cite{gygli14creating}, minimize the reconstruction loss~\cite{mahasseni2017unsupervised}, or directly compute a frame-level score, measuring the frame's contribution in summarizing the video essence~\cite{song2015tvsum}. Various mechanisms and additional cues have been adopted to improve semantics, including temporal dependency~\cite{zhang2016video}, subtitles~\cite{yi2004semantic} and action features~\cite{laganiere2008video}. Although these models are effective in general, they primarily rely on low-level features (\eg, appearance, motion) without much modeling of high-level ``agency'' of human agents.

Obtaining a better semantic summary for videos with rich human interactions and nonverbal communications necessitates the modeling and understanding of the agents' mental world. To tackle this problem, we incorporate belief dynamics and model nonverbal communications in the video summary task for interaction-rich videos.

\section{Representation and Model}

In this section, we start by introducing the proposed \ac{tom} representation, ``five minds,'' that accounts for the triadic relation and ``common mind''; this representation is embedded in a hierarchical graphical model with a six-level structure. Next, to learn a probabilistic distribution over such hierarchically structured data and capture the relations among latent and observable variables, a classic Gibbs energy-based probabilistic formulation with carefully designed and most representative energy terms is derived, capable of parsing the communication events that emerged from the raw pixel inputs and tracking belief dynamics in five minds. At length, we conclude this section with a detailed description of learning and joint inference algorithms.\footnote{Henceforth, we use the term ``mind'' in human and animal studies and the term ``mental state'' in computational models interchangeably.}

\subsection{Hierarchical Representation}\label{sec:mental_representation}

\begin{figure*}
    \centering
    \vspace{-9pt}
    \includegraphics[width=\linewidth]{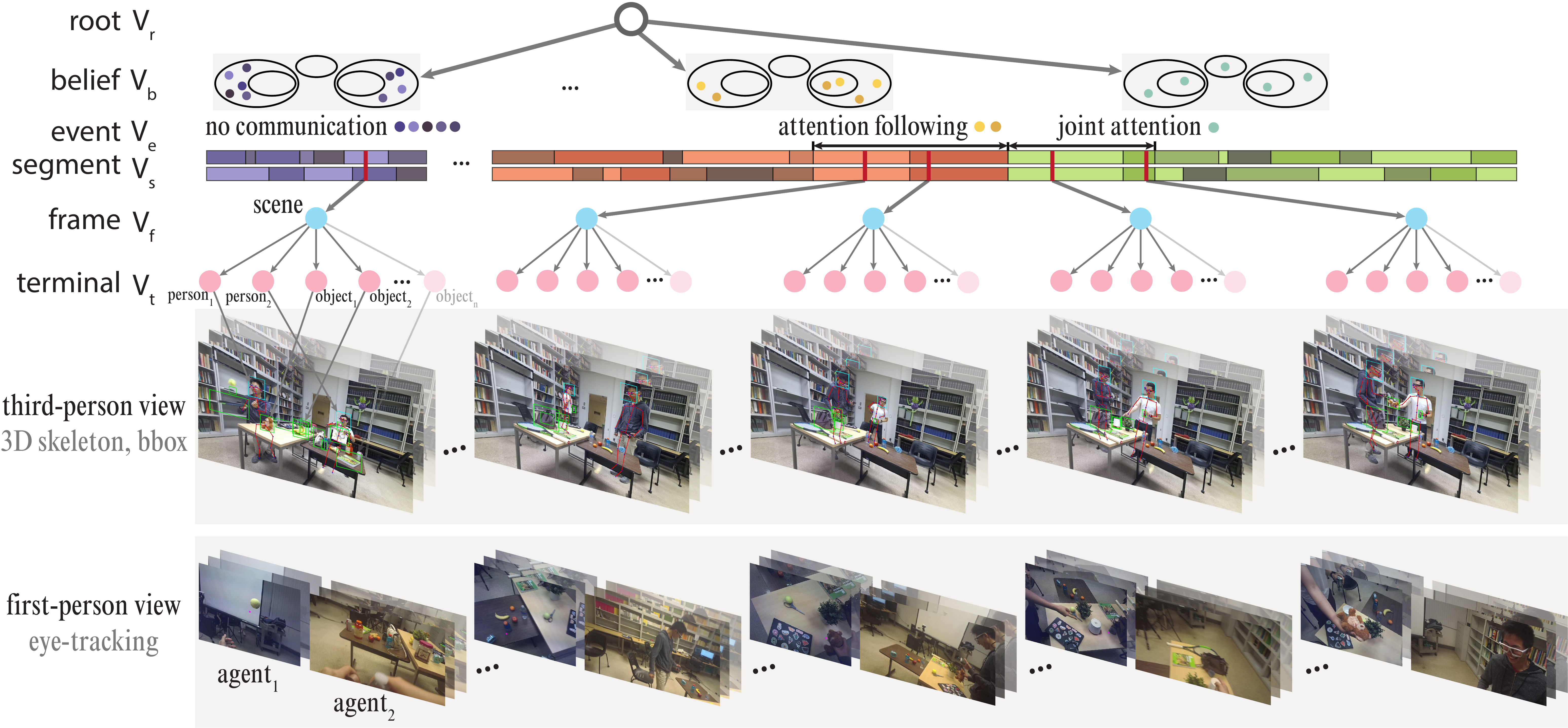}
    \caption{\textbf{A parse graph of a social event with a six-level hierarchical structure.} $V$ denotes vertex sets in the hierarchy. The root node $V_r$ corresponds to the entire video. The set of belief dynamics $V_b$ emerges from the lower-level communication events (see also \cref{fig:concept}). Communication events in $V_e$ decompose into lower-level interactive segments in $V_s$; these segments are social primitives learned unsupervisedly. Each frame of the scene in $V_f$ further decomposes into several terminal nodes in $V_t$, grounded into entities detected from videos. The colored dots in the $V_e$ layer represent belief changes triggered by communication events. Note that belief dynamics are accumulated over time; we only illustrate the most significant changes.}
    \label{fig:pg}
\end{figure*}

Given the input image sequence $I=\{I_t\}_{t=1,...,T}$, the detected human agent $i$ at time $t$ is denoted by $h^i_t=(x^{i}_{t}, p^i_t, g^i_t)$, where $x^{i}_{t} \in \mathbb{R}^3 $ denotes the spatial position, $p^i_t \in \mathbb{R}^{3 \times 26}$ the skeleton pose, and $g^i_t \in \mathbb{R}^3$ the gaze direction. Similarly, $o^j_t=(x^{j}_{t}, c^j_t, d^j_t)$ denotes the detected object $j$ at time $t$, where $x^{j}_{t} \in \mathbb{R}^3$ denotes the spatial location, $c^j_t \in \mathbb{C}$ the object category, and $d^j_t \in \{1,\dots,N_o\}$ the object ID; $\mathbb{C}$ is the object category set. Let $H=\{h^i_t\}$ and $O=\{o^j_t\}$ denote all the detected human agents and objects in the video. Without loss of generality, we assume a minimal setting for triadic relation with two agents in a video.

Formally, all minds $M_t$ at time $t$ is represented as a set, forming a ``five minds'' representation:
\begin{equation}
    M_t=\{m^1_t, m^2_t, m^{12}_t, m^{21}_t, m^c_t\}, \quad{}\quad{} t=1,\dots,T,
\end{equation}
where $m^1_t$ and $m^2_t$ denote two agents' mind, $m^{12}_t$ and $m^{21}_t$ denote the agent's belief about the other agent's mind, and $m^c_t$ denotes their common mind. Each mind is defined as $m_t=\{(o^i_t, A(o^i_t)): i=1, \dots, N_{o,t}\}$ with a set of objects $o^i$ and their attributes $A(o^i)$ (\eg, 3D location). The state change of $M_t$, \ie, $\Delta M_t=\{\Delta m^1_t, \Delta m^2_t, \Delta m^{12}_t, \Delta m^{21}_t, \Delta m^c_t\}$, defines the belief dynamics. Here, $\Delta m=\{\Delta (o^i_t, A(o^i_t))\}$ and belief dynamics in each mind $\Delta (o^i_t, A(o^i_t)) \in \{0, 1, 2, 3\}$, correspond to four communication types, \textit{occur}, \textit{disappear}, \textit{update}, and \textit{null}.

$\Delta M_t$ along time construct the overall \textit{belief dynamics} $\{\Delta \bm{\mathcal{M}}\}$, derived from the spatiotemporal parsing of the video. The parsing is represented by a spatiotemporal parse graph~\cite{zhu2007stochastic} $pg=(pt, E)$, a hierarchical graphical model that combines a parse tree $pt$ and the contextual relation $E$ on terminal nodes; \cref{fig:pg} illustrates an example. A parse tree $pt=(V, R)$ includes the vertex set with a six-level hierarchical structure $V = V_r \cup V_b \cup V_e \cup V_s \cup V_f \cup V_t$ and the decomposing rule $R$, where $V_r$ is the root set with only one node representing the entire video, $V_b$ the set of belief dynamics of ``five minds,'' $V_e$ is the set of communication events, $V_s$ is the set of interactive segments, $V_f$ is the set of frame-based static scenes, and $V_t$ is the set of all the detected instances in an indoor scene. Specifically:%
\setstretch{1}
\begin{itemize}[leftmargin=*,noitemsep,nolistsep]
    \item The belief dynamics $\Delta \bm{\mathcal{M}}$ are conditioned on communication events $V_e$, grouped by interactive segments $V_s$. 
    \item A communication event $e \in V_e$ is one of the three categorical nonverbal communication events: \textit{No Communication}, \textit{Attention Following}, or \textit{Joint Attention}; see \cref{fig:concept}.
    \item An interactive segment $s \in V_s$ is the decomposition of a communication event $e \in V_e$ and represented by the 4D spatiotemporal features $\Phi_s = (\Phi^1_s, \Phi^2_s)$ extracted from detected entities. These features describe social interactions, including both unary $\Phi_s^1$ and pair-wise features $\Phi_s^2$.
    \item The contextual relation $E$ is represented by an attention graph $\mathcal{G}_s$ formed based on 4D features, wherein the node represents an agent or an object in the scene, and an edge is connected between two nodes if there is directed attention detected among the two entities from visual inputs.
\end{itemize}

\subsection{Probabilistic Formulation}

To infer the optimal parse graph $pg^*$ from raw video sequence $I$, we formulate the video parsing of social events as a \ac{map} inference problem:
\begin{equation}
    \small
    \begin{aligned}
        pg^* &=\argmax_{pg} P(pg|H,O)P(H,O|I) \\
        &=\argmax_{pg} P(H,O|pg) P(pg) P(H,O|I),
        \label{eq:posterior}
    \end{aligned}
\end{equation}
where $P(H,O|I)$ is the detection score of agents and objects in the video, $P(pg)$ is the prior model, and $P(H,O|pg)$ is the likelihood model. Below, we detail the prior model and the likelihood model one by one.

\setstretch{0.97}

\paragraph{Prior}

The prior model $P(pg)$ measures the validness of parse graph; all the nodes in the parse graph should be reasonably parsed from the root node. We model the prior probability of $pg$ as a Gibbs distribution: $P(pg) = \frac{1}{Z_1} \exp\{- \energy(pg)\} = \frac{1}{Z_1} \exp \{- \energy_{aggr} - \energy_{evt} - \energy_{be} \}$, where $\energy_{aggr}$ is the aggregation prior, $\energy_{evt}$ the communication event prior, and $\energy_{be}$ the belief dynamics prior. Specifically,
\begin{itemize}[leftmargin=*,noitemsep,nolistsep]
    \item The aggregation prior is defined as $\energy_{aggr}=\lambda_1 \frac{N_e}{T}$ to encourage the algorithm to focus more on high-level communication patterns, instead of being trapped into trivial primitives that results in fragmented segmentation.
    \item The communication event prior leverages transition and co-occurrence frequencies of communication events,
    \begin{equation}
        \small
        \begin{aligned}
            \energy_{evt} = & -\frac{\lambda_2\sum_{i, j, \mathbbm{1}^{trans}(e_i, e_j) = 1} \log p^{trans}(e_i, e_j)}{\sum_{i, j} (\mathbbm{1}^{trans}(e_i, e_j) = 1)} \\
            & - \frac{\lambda_3\sum_{i, j, \mathbbm{1}^{occ}(e_i,e_j) = 1} \log p^{occ}(e_i, e_j)}{\sum_{i, j} (\mathbbm{1}^{occ}(e_i, e_j) = 1)},
            \label{eq:prior_event}
        \end{aligned}
    \end{equation}
    where $p^{trans}(e_i, e_j)$ and $p^{occ}(e_i, e_j)$ are based on frequencies from the dataset, and $\mathbbm{1}^{trans}$ and $\mathbbm{1}^{occ}$ are indicator functions that reflects the spatiotemporal relations.
    
    \item $\energy_{be}$ models the prior of belief dynamics, which helps to prune some invalid configurations, such as two consecutive \textit{occur}s or an \textit{occur} after an \textit{update}. The prior model is defined as 
    $\energy_{be}=-\lambda_4\sum_{j=1}^{N_e}\log p^M(\Delta \bm{\mathcal{M}}_j | e_j)$, and
    \begin{equation}
        \small
        p^M(\Delta \bm{\mathcal{M}}_j | e_j) = \prod_t p(\Delta M_{t+1} | \Delta M_t, e_j) \, p(\Delta M_t | e_j),
        \label{eq:prior_mind}
    \end{equation}
    where $ \Delta \bm{ \mathcal{M}}_j$ is the set of belief dynamics in event $e_j$.
\end{itemize}

\setstretch{1}

\paragraph{Likelihood}

The likelihood model measures the consistency between the parse graph $pg$ and the ground-truth observed data $H$ and $O$. Since our model has a hierarchical structure, we split the likelihood into three energy terms, corresponding to the three crucial layers in the parse graph:
\begin{equation}
    \small
    \begin{aligned}
        & P(H, O|pg) = P(H, O| V_b, V_e, E) \\
        & =  \frac{1}{Z_2} \exp\{- \energy^{comp}(H, O| V_e, E) \\
        & - \energy^{evt}(H, O| V_e, E) - \energy^{be}(H, O, V_e| \{\Delta \bm{\mathcal{M}}\})\}.
        \label{eq:likelihood_full}
    \end{aligned}
\end{equation}

\begin{itemize}[leftmargin=*,noitemsep,nolistsep]
    \item The first energy term $\energy^{comp}$ constrains the communication event composed by the interactive segments, so that the features within one composition are sufficiently similar, whereas the features between two consecutive compositions are considerably distinct:
    \begin{equation}
        \small
        \begin{aligned}
            & \energy^{comp}(H, O| V_e, E) =  \energy( \bm\varPhi | V_s, E) \\ 
            & =\frac{\lambda_5}{N_e} \sum^{N_e}_{j=1} \left( \frac{1}{T_j} \sum_t \mathcal{D}( \phi_{{j},t}, \phi_{{j}, t+1} ) \right)\\
            & - \frac{\lambda_6 \sum_{i, j, \mathbbm{1}^{trans}(e_i, e_j) = 1} \mathcal{D}(\psi(\Phi_i), \psi(\Phi_j))}{\sum_{i, j} (\mathbbm{1}^{trans}(e_i, e_j) = 1)}  \\ 
            & - \frac{\lambda_7 \sum_{i, j, \mathbbm{1}^{occ}(e_i, e_j) = 1} \mathcal{D}(\psi(\Phi_i), \psi(\Phi_j))}{\sum_{i, j} (\mathbbm{1}^{occ}(e_i, e_j) = 1)}
        \end{aligned}
        \label{eq:likelihood_segment}
    \end{equation}
    where $\Phi_i = \{ \phi_{i, t} \}$ is the set of features within the interactive segment $s_i$, $\psi(\cdot)$ the wavelet transform~\cite{quennesson2006wavelet}, and $\mathcal{D}(\cdot)$ the Euclidean distance between the two sets of features.

    \item The second energy term $\energy^{evt}$ is the negative communication event classification score with respect to the detected feature set $\bm{\varPhi} = \{ \Phi_j \}$ and the constructed attention graph set $\bm{\mathcal{G}} = \{ \mathcal{G}_j \}$. This second term is defined as $\energy^{evt}(H, O|V_e, E) = \energy( \bm\varPhi, \bm{\mathcal{G}} | V_e)$ and encodes all the entities in the scene extracted from visual input, which can be solved by a \ac{mle}:
    \begin{equation}
        \small
        \begin{aligned}
            \energy( \bm\varPhi, \bm{\mathcal{G}} | V_e) & = - \frac{1}{N_e} \sum^{N_e}_{j = 1} \lambda_8 \log p(\Phi_{\Lambda_{j}}, \mathcal{G}_{\Lambda_j} | e_j) \\
            &= - \frac{1}{N_e} \sum^{N_e}_{j = 1} \lambda_8 \log p(e_j | \Phi_{\Lambda_j}, \mathcal{G}_{\Lambda_j}) - C,
            \label{eq:likelihood_event}
        \end{aligned}
    \end{equation}
    where $\Lambda_j$ is the set of indexes of the interactive segments decomposed from $e_j$, and $C$ is a constant.

    \item The third energy term $\energy^{be}$ models the belief dynamics in all five minds, formally defined as
    \begin{equation}
        \scalebox{0.82}{$
        \begin{aligned}
            &\energy^{be}(H, O, V_e| \{\Delta \bm{\mathcal{M}}_j \}) = - \frac{1}{N_e} \sum^{N_e}_{j = 1} \lambda_9 \log p( \Delta \bm{ \mathcal{M}}_j  | H, O,
             V_e) \\ & = -\frac{1}{N_e} \sum^{N_e}_{j=1} ( \frac{1}{T_j} \sum_t \lambda_9 \log p (\Delta M_{j,t+1} | \mathit{g}_{j,t+1}, e_j,
             \{ \Delta M_{j, t'} \} )),
            \label{eq:likelihood_mind}
        \end{aligned}
        $}
    \end{equation}
    where $t' \in [t_j^s, t]$, $t_j^s$ is the starting frame of the event $e_j$, and $\mathit{g}_{j,t+1}$ the attention graph of frame $t+1$ in event $e_j$.
\end{itemize}

\setstretch{0.97}

\subsection{Learning}

The detailed learning process follows a bottom-up procedure. Specifically, the algorithm (i) parses each frame to extract the entities and relations, (ii) jointly and dynamically parses both interactive segments (proposals generated unsupervisedly by clustering methods) and communication events (with trained likelihood) by beam search, (iii) predicts the belief dynamics (with trained likelihood), and (iv) fine-tunes all the parameters to minimize the overall loss. \cref{alg:learning} details the overall learning procedure.

\begin{algorithm}[ht!]
    \fontsize{8}{10}\selectfont
    \caption{Learning to parse social events}
    \label{alg:learning}
    \SetAlgoLined
    \DontPrintSemicolon
    \SetKwInOut{Input}{Input}
    \SetKwInOut{Output}{Output}
    \SetKwInOut{Initialization}{Init.}
 
    \Input{Video $\{I_{train}\}$, ground truth $V_e^*$ and $V_b^*$.}
    \Output{Parameter sets $\Theta_1^*$ and $\Theta_2^*$, and parse graph $pg$.}
    \Initialization{$H,O,\bm\varPhi,\bm{\mathcal{G}},\Theta_1^*,\Theta_2^*=\emptyset{}$; $L_1^*,L_2^*=+\infty$}
    
    \For{$I_i$ in $\{I_{train}\}$}{
        $H_i$ = \mbox{humanDetectionWithReID($I_i$)}, $H \xleftarrow{} H \cup H_i$ \;
        $O_i$ = \mbox{objectDetectionWithReID($I_i$)}, $O \xleftarrow{} O \cup O_i$ \;
        $\bm\varPhi_i$ = \mbox{extractSTFeatures($H_i$, $O_i$)}, $\bm\varPhi \xleftarrow{} \bm\varPhi \cup \bm\varPhi_i $ \;
        $\bm{\mathcal{G}}_i$ = \mbox{buildAttentionGraph($H_i$, $O_i$, $\bm\varPhi_i$)}, $\bm{\mathcal{G}} \xleftarrow{} \bm{\mathcal{G}} \cup \bm{\mathcal{G}}_i$ \;
    }
    $V_s \xleftarrow{}$ Generate $\{ s \}$ by unsupervised clustering. \;
    \tcc{Train likelihood of $e_j$ as in~\cite{fan2019understanding}}
    Train $p(e_j | \Phi_{\Lambda_j}, \mathcal{G}_{\Lambda_j})$ in \cref{eq:likelihood_event} with ground-truth $V_e^*$. \;
    \tcc{Finetune the parameter set $\Theta_1^*$.}
    \For{$\Theta^{(i)}_1 = (\lambda_1, \lambda_2, \lambda_3, \lambda_5, \lambda_6, \lambda_7, \lambda_8)$ $\in$ $\Omega_{\Theta_1}$}{
        Compute $\energy^{comp}$ based on \cref{eq:likelihood_segment}, given $\bm\varPhi$ and $\Theta^{(i)}_1$. \;
        Compute $\energy^{evt}$ based on \cref{eq:likelihood_event}, given $\bm\varPhi$, $\bm{\mathcal{G}}$, $\Theta^{(i)}_1$. \;
        Infer $V_e$ by dynamic programming beam search; see details in \cref{alg:beam_search}. \;
        Calculate error $L_1$ between $V_e$ and $V_e^*$. \;
        \lIf{$L_1 < L_1^*$} { $L_1^* \xleftarrow{} L_1$. $\Theta_1^* \xleftarrow{} \Theta^{(i)}_1$.
        }
    }
    \tcc{Train belief dynamics likelihood}
    Train $p\left(\Delta M_{j, t+1} | \mathit{g}_{j,t+1}, e_j, \{ \Delta M_{j, t'} \} \right)$ in \cref{eq:likelihood_mind} with $V_b^*$. \;
    \tcc{Finetune the parameter set $\Theta_2^*$.}
    \For{$\Theta^{(i)}_2$ = $(\lambda_4, \lambda_9)$ $\in$ $\Omega_{\Theta_2}$}{
        \For{$e_j$ in $V_e$} {
            Compute the posterior probability of belief dynamics based on \cref{eq:prior_mind,eq:likelihood_mind}. \;Predict the best $\hat{V_b}$ by MAP. \;}Calculate error $L_2$ between the best predicted belief dynamics $\hat{V_b}$ and the ground-truth $V_b^*$. \;
        \lIf{$L_2 < L_2^*$} {$L_2^* \xleftarrow{} L_2$. $\Theta_2^* \xleftarrow{} \Theta^{(i)}_2$. 
        }
    }
\end{algorithm}

\begin{algorithm}[ht!]
    \fontsize{8}{10}\selectfont
	\caption{Event inference via DP beam search}
	\label{alg:beam_search}
	\SetAlgoLined
	\DontPrintSemicolon
 
	\SetKwInOut{Input}{Input}\SetKwInOut{Output}{Output}
	\SetKwInOut{Initialization}{Initialization}
 
	\Input{$\bm\varPhi$, $\bm{\mathcal{G}}$, $V_s$, $p(e_j | \Phi_{\Lambda_j}, \mathcal{G}_{\Lambda_j})$.}
	\Output{$V_e$}%
	
	\Initialization{$V_e=\emptyset{}, \mathcal{B}=\{V_e, p=0\}, m,n.$}
    \While{True}{ 
        $\mathcal{B'}=\emptyset{}$ \;
        \For{$\{ V_e, p \} \in \mathcal{B}$}{
            $\{e_i\}=Next(V_s, V_e,m)$ \;
            \leIf{$\{e_i\}$ $\neq$ $\emptyset{}$}{ \;
                \For{each proposed $e_i$}{
                    $p(V_e |\bm\varPhi, \bm{\mathcal{G}})=DP(V_e, p, e_i, \Phi, \mathcal{G})$ \;
                    $V_e=V_e \cup \{e_i\}$;
                    $\mathcal{B'}=\mathcal{B'} \cup \{V_e, p\}$ \;
                    
                }
            }{$\mathcal{B'}=\mathcal{B'} \cup \{V_e,p\}$}
        }
       
        \leIf{$\mathcal{B'} == \mathcal{B}$}{
        return $V_e= Best(\mathcal{B}, 1)$ \;
        }{
             $\mathcal{D}=Best(\mathcal{B'}, n)$;
            $\mathcal{B} = \mathcal{D}$ 
        }
    }
\end{algorithm}

\vspace{-6pt}
\section{Experiment}

\begin{table*}[t!]
    \centering
    \vspace{-9pt}
    \caption{\textbf{Statistics of belief dynamics in our dataset.} The numbers of belief dynamics denote different categories: 0--\textit{occur}, 1--\textit{disappear}, 2--\textit{update}, 3--\textit{null}. The belief dynamics are imbalanced by its inherent sparse nature, with \textit{null} most frequent and \textit{occur}/\textit{disappear} rare; it is one of the many challenges that make the inference of belief dynamics difficult.
    }
    \label{tab:statistics}
    \resizebox{\linewidth}{!}{
        \begin{tabular}{r aaaa bbbb aaaa bbbb aaaa}
            \toprule
            \multirow{-1}{*}{Five Minds}&\multicolumn{4}{c}{$m^1$}&\multicolumn{4}{c}{\textit{$m^2$}}
            &\multicolumn{4}{c}{\textit{$m^{12}$}}&\multicolumn{4}{c}{\textit{$m^{21}$}}&\multicolumn{4}{c}{\textit{$m^c$}}\\
            \hline
            Belief Dynamics &0 &1 &2 &3 &0 &1 &2 &3 &0 &1 &2 &3 &0 &1 &2 &3 &0 &1 &2 &3\\
            \hline
            No Communication &49 &0 &501 &78017 &36 &2 &442 &78087 &1 &1 &0 &78514 &3 &1 &0 &78528 &0 &0 &0 &78545\\
            Attention Following &36 &6 &401 &36953 &33 &2 &457 &36899 &23 &6 &128 &37238 &23 &5 &264 &37104 &0 &0 &0 &37370\\
            Joint Attention & 15 &5 &324 &26136 &17 &0 &340 &26119 &32 &6 &290 &26153 &28 &1 &267 &26180 &32 &6 &166 &26276\\
            \bottomrule
        \end{tabular}
    }
\end{table*}

\begin{table*}[t!]
    \centering
    \caption{\textbf{Quantitative results on predicting belief dynamics of five minds.} The best scores are marked in \textbf{bold}.}
    \resizebox{\textwidth}{!}{%
        \begin{tabular}{r|cccccc|cccccc}
            \bottomrule
            \rowcolor{mygray}
            & 
            \multicolumn{6}{c}{Macro-average of Precision ($\uparrow$)} & \multicolumn{6}{c}{Macro-average of F1-score ($\uparrow$)} \\
            \hline
            Model & $m^1$ & $m^2$ & $m^{12}$ & $m^{21}$ & $m^{c}$ & Avg. & $m^{1}$ & $m^{2}$ & $m^{12}$ & $m^{21}$ & $m^{c}$ & Avg. \\
            \hline\hline
            Chance & 0.250 & 0.250 & 0.250 & 0.250 & 0.250 & 0.250 & 0.103 & 0.104 & 0.102 & 0.101 & 0.100 & 0.102 \\
            CNN & 0.250 & 0.250 & 0.250 & 0.250 & 0.250 & 0.250 & 0.171 & 0.167 & 0.169 & 0.174 & 0.250 & 0.186 \\
            CNN w/ HOG-LSTM & 0.250 & 0.250 & 0.250 & 0.250 & 0.250 & 0.250 & 0.167 & 0.132 & 0.205 & 0.182 & 0.250 & 0.187\\
            CNN w/ HOG \& memory & 0.277 & 0.279 & 0.266 & 0.267 & 0.259 & 0.270 & 0.285 & 0.285 & 0.246 & 0.250 & 0.155 & 0.244\\
            Features w/ memory & 0.272 & 0.278 & 0.253 & 0.260 & 0.256 & 0.264 & 0.274 & 0.288 & 0.230 & 0.227 & 0.191 & 0.242\\
            \hline
            Init. seg. & 0.371 & \textbf{0.418} & 0.293 & 0.301 & 0.265 & 0.330 & 0.346 & 0.366 & 0.302 & 0.314 & 0.274 & 0.320\\
            Event prior & 0.384 & 0.409 & 0.294 & 0.307 & 0.264 & 0.332 & 0.365 & 0.364 & 0.305 & 0.324 & 0.273 & 0.326\\
            Uniform event & 0.385 & 0.413 & 0.293 & 0.310 & 0.267 & 0.334 & 0.363 & 0.366 & 0.304 & 0.328 & 0.278 & 0.328\\
            \hline
            \textbf{Ours-full} & \textbf{0.397} & 0.415 & \textbf{0.316} & \textbf{0.315} & \textbf{0.278} & \textbf{0.344} &\textbf{0.431} & \textbf{0.443} & \textbf{0.351} & \textbf{0.349} & \textbf{0.299} & \textbf{0.375}\\
            \toprule
        \end{tabular}%
    }%
    \label{tab:performance}
\end{table*}

\subsection{Dataset}

To verify the efficacy of the proposed algorithm, we collected a new 3D video dataset with rich social interactions. This dataset was shot in both third-person view (with Azure Kinect sensors) and first-person view (with two pairs of glasses, capable of reading videos) to properly mimic human social interactions and the perception of the environment. One pair of glasses is an SMI eye tracker, capable of accurately tracking human eye gazes while recording the video. Another pair is Pivothead glasses, simply recording the first-person view video. Both glasses have a similar look to regular glasses to ensure maximum naturalism during data collection. Crucially, such a setup also eases the challenging ground-truth annotation procedure; the annotators can better understand the belief dynamics with the precise reference of the agents' first-person view attention.

Our dataset is specially designed to cover typical nonverbal communication in rich social interactions. In total, we collected 88 videos (109,331 frames, 72.89 minutes of each sensor) recorded with 12 participants in 7 different scenarios. The participants were asked to perform three types of nonverbal communication naturally, as illustrated in \cref{fig:concept}. Among all the nonverbal behaviors in the dataset, we annotated two major types including eye gaze (almost all frames) and pointing (around 5.47\% of all frames); note also that our scenarios involve both first-order and second-order false beliefs~\cite{nematzadeh2018evaluating}. We did not provide scripts for performing detailed actions; instead, we only informed the participants about the task and the types of nonverbal communication they can use. This design follows the principles in recent large-scale video dataset collection~\cite{jia2020lemma} to ensure maximum realism.

We densely annotated the dataset, including human head bounding boxes, object bounding boxes, pointing, interactive segments, communication events, and the belief dynamics in all ``five minds'' for all the objects in all the scenes. These ground-truths were first annotated by seven volunteers using Vatic~\cite{vondrick2013efficiently} after simple tutorials and later verified by at least one expert. The annotation process relied on synced third-person and first-person views. Of note, having perfect ground-truth annotation is impossible for any high-level semantic task (\eg, belief dynamics) due to its intrinsic ambiguities, which have also been exhibited in other more traditional computer vision tasks (\eg, activity recognition and event segmentation). Here, the goal of annotation is not to provide universally perfect labels at each frame; instead, we hope to use these annotations to provide a reasonable quantitative measurement of the model's performance. In total, 5,975 frames are labeled with pointing gestures. Among communication events, $48.56\%$ is \textit{No Communication}, $32.51\%$ \textit{Attention Following}, and $18.93\%$ \textit{Joint Attention}. 40 videos have first-order false beliefs, and 13 videos have second-order false beliefs. 26 videos are reserved as the testing set, and the rest 62 videos are used for model training and validation. Detailed statistics of belief dynamics in the dataset are tabulated in \cref{tab:statistics}.

The pre-process procedure including following steps:
\begin{itemize}[leftmargin=*,noitemsep,nolistsep]
    \item Azure Kinect SDK tracks 26 3D joints of each agent.
    \item Detectron2~\cite{wu2019detectron2} detects objects.
    \item Deepsort~\cite{wojke2018deep} tracks objects. 
    \item Object RGB and category features are matched and aligned between the third- and first-person views.
    \item Gaze360~\cite{xu2018gaze} estimates 3D human gazes.
    \item We use the detected objects, depth maps, and camera parameters to recover 3D object point clouds.
\end{itemize}
Combining with 3D information and multi-views, features that can be potentially extracted from our dataset would be advantageous compared to 2D data, especially in cases that require handling complex occlusions, which is crucial for multi-agent human-object and human-human interactions.

Although collecting and processing a new 3D video dataset is challenging, it is the only viable direction to go if we hope to study social cognition on natural videos in indoor environments. First, as such a study requires dense annotations for evaluations, it demands specific hardware and computational power (\eg, eye-tracking glasses, Azure Kinect sensors) to collect the raw data in a way that could ease the annotation process. Hence, crowd-sourcing the dataset is not an option. Second, the ideal dataset would cover rich social interactions with nonverbal communication. The closest existing dataset is presented in Fan \etal~\cite{fan2019understanding} using clips collected from TV shows and movies; however, its nonverbal communication and belief dynamics are sparse. As a significant upgrade to existing datasets, we hope this new dataset paves the way towards \ac{tom} modeling in natural and complex indoor environments.

\subsection{Task 1: Predicting Belief Dynamics} \label{sec:task_1}

To directly evaluate the proposed algorithm, we predict belief dynamics in all five minds on our dataset and evaluate based on the Macro-average of Precision and F1-score.

To make a fair comparison, we consider the following five \textbf{baselines}:
(i) \textbf{Chance} is a weak baseline, \ie, randomly assigning a belief dynamic label;
(ii) \textbf{CNN} uses the pre-trained ResNet-50~\cite{he2016deep} to extract image features and adopts an MLP to classify the belief dynamics;
(iii) \textbf{CNN w/ HOG-LSTM} feeds both the ResNet-50 features of the entire image and the HOG~\cite{dalal2005histograms} of the local image patch gazed by the agent to an LSTM~\cite{hochreiter1997long}, followed by an MLP to predict belief dynamics;
(iv) \textbf{CNN w/ HOG \& memory} adds the history of predicted belief dynamics on top of ResNet-50 and HOG features;
(v) \textbf{Features w/ memory} uses the same sets of features as our methods (see details below) with the history of predicted belief dynamics; all features are concatenated and fed into an MLP to predict belief dynamics. We only used the annotations of belief dynamics in all five minds when training the deep learning models.

To assess the contributions and efficacy of essential components in the proposed method, we derive the following variants as \textbf{ablation study}:
(i) \textbf{Init. seg.} directly uses the initial interactive segments generated by unsupervised clustering as event segments, without additional temporal clustering by beam search;
(ii) \textbf{Event prior} only uses event prior for event assignment without the event likelihood; 
(iii) \textbf{Uniform event} replaces event posterior with uniform distribution, and randomly assign event labels for all segments.

A suite of 4D spatiotemporal features $\Phi_s$ (see \cref{sec:mental_representation}) are adopted to ground our methods to raw video inputs. Unary feature $\Phi^1_s$ concerns a single agent and concatenates features of human poses, hand-object distances, and estimated gaze and attention. Pair-wise feature $\Phi^2_s$ focuses on the relations between two agents and concatenates features of the relative human poses between two agents, relative gaze angles, and relative hand joint distances. All models are implemented in PyTorch using ADAM optimizer~\cite{kingma2014adam} and trained on an Nvidia TITAN RTX GPU.

Quantitatively, our full model achieves the best performance on predicting belief dynamics of five minds, measured by the macro-average of both Precision and F1-score; see comparisons in \cref{tab:performance}. Overall, the CNN baseline and its variant with HOG-LSTM perform the worst on this challenging task, only slightly better than randomly guessing. The performances of \textit{CNN w/ HOG \& memory} and \textit{Features w/ memory} are improved after incorporating the history of the estimated belief dynamics. The results indicate that the performance bottleneck to infer belief dynamics does not lie in the low-level features or representations; instead, it heavily depends on high-level semantics to distinguish similar segments and events to predict mental states and belief dynamics in nonverbal communication correctly.

The ablation study further reveals the effects of various model components. Without temporal segment re-clustering by dynamic-programming-based beam search, \textit{Init. seg.} performs worse. Compared to using the posterior event distribution in our full model, the performance would drop if using either the biased event distribution prior or the uniform event distribution. Our full model yields the best performance on this challenging task with carefully derived hierarchical representation and joint learning algorithms.

\begin{figure*}[t!]
    \centering
    \vspace{-9pt}
    \includegraphics[width=\linewidth]{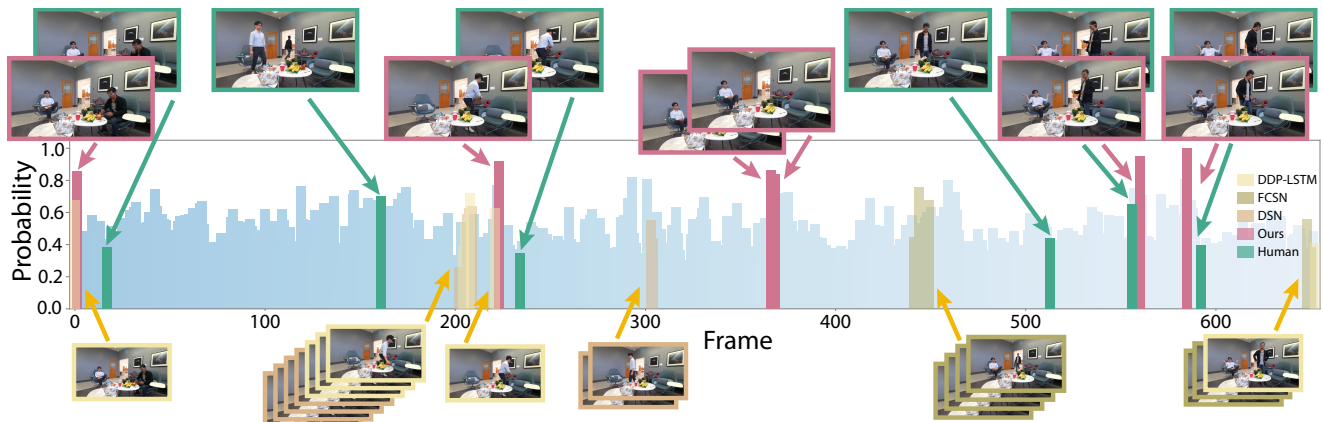}
    \caption{\textbf{Qualitative comparisons of keyframe-based video summarization.} The blue histogram represents the estimated probability of belief dynamics (including \textit{occur} and \textit{disappear}) by our model. The top keyframes chosen by human participants are shown next to our model's prediction. Overall, baseline models tend to predict frames merely based on visual patterns, making the top keyframes similar to each other and grouped in clusters. In comparison, with the proper modeling of belief dynamics, our method tends to capture the moment with significant belief changes in ``five minds,'' resulting in a more uniform set of keyframes along the time. Note that human participants also demonstrate similar behaviors when choosing the top keyframes for video summarization. Please refer to the \textit{supplementary material} for additional qualitative results.}
    \label{fig:keyframe}
\end{figure*}

\subsection{Task 2: Keyframe-based Video Summary}

\begin{figure}[ht!]
    \centering
    \includegraphics[width=\linewidth]{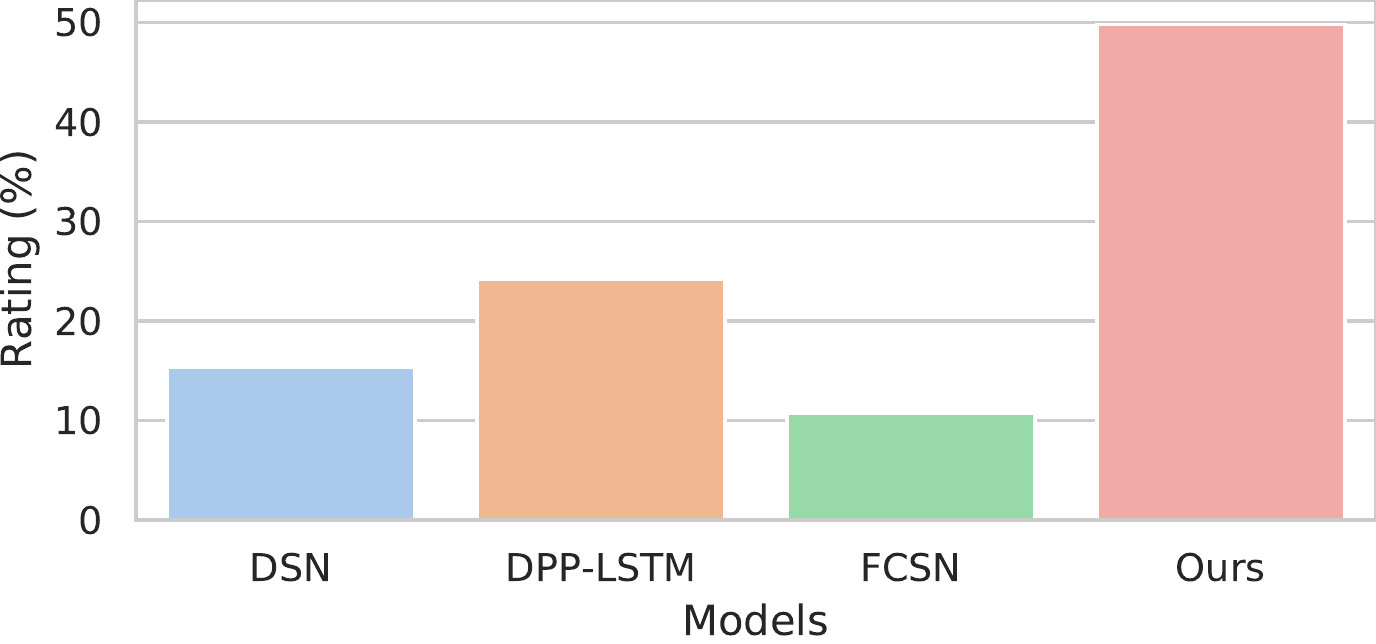}
    \caption{\textbf{Human ratings of keyframe-based video summary.} Our model outperforms state-of-the-art methods significantly on videos with rich interactions.}
    \label{fig:rating}
\end{figure}

We apply the proposed method on keyframe-based video summary on videos with rich social interactions. For comparison, we choose three state-of-the-art methods as baselines: DPP-LSTM~\cite{zhang2016video}, FCSN~\cite{rochan2018video}, and DSN~\cite{zhou2018deep}. To adopt the proposed method for this task, we sum over the predicted probabilities of belief dynamics \textit{occur} and \textit{disappear}, which indicate a significant belief change in agents' minds; we use it as the score indicating the frame's contribution in summarizing the video content. To quantitatively compare our method with three baselines, we conducted a study with 33 human participants, who were neither experts on the task, nor were they knowledgeable about the video. After seeing the entire video, participants were presented with top keyframes from different methods (in a counterbalanced order to avoid bias) and asked to select the best keyframe-based summary that describes the observed video. The proposed method outperforms the state-of-the-art methods significantly; see the detailed comparisons in \cref{fig:rating}.

\setstretch{0.99}

We further discuss a qualitative comparison shown in \cref{fig:keyframe}; please refer to the \textit{supplementary material} for additional qualitative comparisons. In this example, agent A (in a black jacket) puts a teddy bear on the desk and leaves the room. Agent B (in a white shirt) later hides the teddy bear into a bag. When agent A comes back, he cannot find the teddy bear at the original location, showing his confusion to agent B by spreading his hands. Agent B pretends to have no clue about what happened and also spreads his hands. Agent A looks around for the lost teddy bear helplessly. As shown on the top of \cref{fig:keyframe} (in green), keyframes chosen by human participants give a complete and refined summary of the video content. Our model produces a similar story digest---the most similar one compared with human judgments among all the methods. In essence, our method captures almost all the crucial moments of the story by modeling belief dynamics during social interactions. The keyframes generated by other baseline methods fail to capture the key moments spanning across the entire story; they tend to group the predicted keyframes in selected moments.

Taken together, the result presented here is no surprise. When watching and summarizing the video with rich social interactions, humans primarily understand the story from a higher level, typically considering the issues going on in the mental world instead of purely looking at the visual motions. As such, by introducing higher-level multi-agent belief dynamics into the keyframe modeling and selecting procedure, the generated keyframes can be mostly optimized to understand social interactions better.
 
\section{Conclusion}
 
This paper studies two critical components in understanding multi-agent social interaction in 3D real scenes, \ie, understanding nonverbal communication and belief dynamics in ``five minds,'' with a particular focus on a structured mental representation of ``common mind.'' A six-level hierarchical graphical model is devised to account for the parsing of belief dynamics in ``five minds,'' nonverbal communication events, the 4D spatiotemporal interactive segments, and the detected entities and relations from raw visual inputs. We propose an energy-based probabilistic model and a beam-search-based algorithm to learn and infer communication events and belief dynamics jointly. Experimental results show that our model captures the sparse belief dynamics in all five minds and facilitates generating more comprehensive keyframe-based video summarization. We believe such a unique social aspect of scene understanding could have broader applications in various future tasks.

\paragraph{Acknowledgements}

The authors thank Baoxiong Jia at UCLA and Siyuan Qi at Google Inc. for helpful discussions. Tao Gao was supported in part by DARPA PA 19-03-01 and ONR MURI N00014-16-1-2007. Other authors were supported in part by ONR MURI N00014-16-1-2007, ONR N00014-19-1-2153, and DARPA XAI N66001-17-2-4029.

{\small
\setstretch{1}
\bibliographystyle{ieee_fullname}
\bibliography{reference}
}

\end{document}